\documentclass{article}
\usepackage{spconf,amsmath,graphicx}
\usepackage{booktabs}
\usepackage{makecell}
\usepackage{color}
\usepackage{amssymb}
\usepackage[table]{xcolor}

\newcommand{\ourmethod}{\textsc{u-EIDG}}
\newcommand{\task}{KIDG}

\title{Unleashing Potential of Evidence in Knowledge-Intensive \\ Dialogue Generation}
\name{
Xianjie Wu\textsuperscript{\rm 1}, Jian Yang\textsuperscript{\rm 1 *\thanks{* Corresponding Author}}, Tongliang Li\textsuperscript{\rm 1 *}, Di Liang\textsuperscript{\rm 2}, Shiwei Zhang\textsuperscript{\rm 3}, Yiyang Du\textsuperscript{\rm 1}, Zhoujun Li\textsuperscript{\rm 1}
}
\address{
  \textsuperscript{\rm 1}Beihang University,  
  \textsuperscript{\rm 2}Fudan University, \textsuperscript{\rm 3}Baidu Inc \\
  \{wuxianjie, jiaya, tonyliangli, duyiyang, lizj\}@buaa.edu.cn, \\
  \{liangd17\}@fudan.edu.cn, \{zhangshiwei05\}@baidu.com
}

\begin{document}
\maketitle

\begin{abstract}
Incorporating external knowledge into dialogue generation (\task{}) is crucial for improving the correctness of response, where evidence fragments serve as knowledgeable snippets supporting the factual dialogue replies. 
However, introducing irrelevant content often adversely impacts reply quality and easily leads to hallucinated responses. Prior work on evidence retrieval and integration in dialogue systems falls short of fully leveraging existing evidence since the model fails to locate useful fragments accurately and overlooks hidden evidence labels within the KIDG dataset.
To fully \textbf{U}nleash the potential of evidence, we propose a framework to effectively incorporate \textbf{E}vidence in knowledge-\textbf{I}ntensive \textbf{D}ialogue \textbf{G}eneration (\ourmethod{}). 
Specifically, we introduce an automatic evidence generation framework that harnesses the power of Large Language Models (LLMs) to mine reliable evidence veracity labels from unlabeled data. By utilizing these evidence labels, we train a reliable evidence indicator to effectively identify relevant evidence from retrieved passages. 
Furthermore, we propose an evidence-augmented generator with an evidence-focused attention mechanism, which allows the model to concentrate on evidenced segments. Experimental results on MultiDoc2Dial demonstrate the efficacy of evidential label augmentation and refined attention mechanisms in improving model performance. Further analysis confirms that the proposed method outperforms other baselines (+3$\sim$5 points) regarding coherence and factual consistency.


\end{abstract}
%



\section{Introduction}
\label{sec:introduction}

\begin{figure}[htb]
\centering
\includegraphics[width=0.45\textwidth]{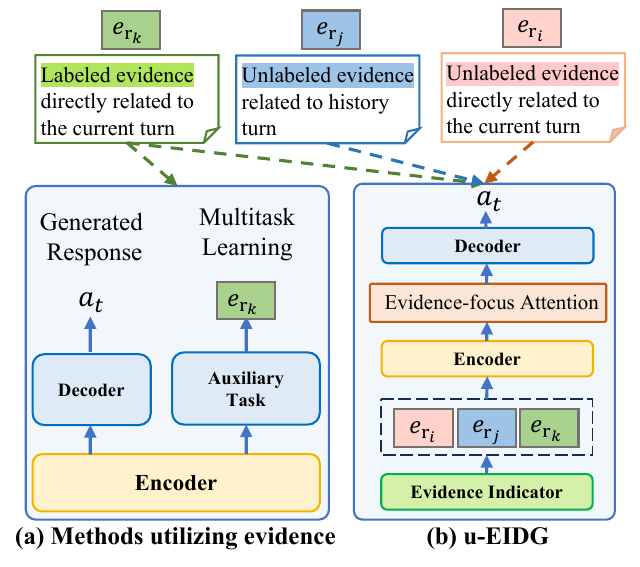}
\caption{Comparison between the multi-task baseline and our method: \textcolor{green}{$e_{r_{k}}$}, \textcolor{blue}{$e_{r_{j}}$}, and \textcolor{red}{$e_{r_{i}}$}, each explained specifically in the diagram. Our approach outperforms other methods by effectively leveraging the potential of unlabeled evidence.}
\label{fig:intro-case}
\vspace{-15pt}
\end{figure}
Knowledge-intensive dialogue generation (KIDG) \cite{dinan2018wizard,feng2021multidoc2dial,li-etal-2022-knowledge,li-etal-2022-enhancing-knowledge,kinet,griprank} facilitates multi-turn conversational interactions by retrieving domain-specific documents. 
Compared to the open-domain question-answering task \cite{voorhees-tice-2000-trec,joshi-etal-2017-triviaqa, SachanRHDY21}, KIDG allows for more specialized dialogues, specifically in domains like biomedicine and law.
Some researchers \cite{chen-etal-2017-reading,chen-etal-2020-bridging} try to fetch Wikipedia documents and extract answers from the corresponding documents for knowledge-driven dialogue. Further, retrieval augmented generation \cite{lewis2020retrieval,karpukhin2020dense,alm,ganlm,izacard2021leveraging,glass2022re2g,geigle2022retrieve,izacard2021distilling} adopts a dense passage retriever (DPR) for knowledge selection and combines it with a seq2seq generation style, where the retrieved fragments are fused in the encoder or decoder.

Under the scenario of the KIDG sensitive to the retrieved segments, more accurate matching between the question and the external documents is required \cite{feng2021multidoc2dial}. The irrelevant passages may bring some noise and redundant content, increasing the difficulty for the model to choose the proper information to support the knowledge-driven dialogue \cite{zheng-etal-2023-contextual}. Besides, the existing KIDG benchmarks only annotate a single correct evidence passage per query turn and ignore the potential evidence in other passages \cite{dinan2018wizard,feng2021multidoc2dial}, which can not provide sufficient support for dialogue due to the sparse evidence annotation. In Fig. \ref{fig:intro-case}, previous evidence-based methods tend to construct implicit auxiliary evidence-related training objectives instead of directly modeling evidence. The baseline \cite{wu-etal-2021-dialki,asai2022evidentiality,zhang2023coarse} optimizes the joint objective of the response generation and evidence identification, which predicts the response and determines the evidentiary support for the response. The joint modeling approach fails to utilize informative evidence content directly and overlooks the potential evidence within the passage. \textit{How to explicitly leverage the external evidence still requires further exploration.}

To fully \textbf{U}nleash the potential of evidence, we propose a framework to effectively incorporate \textbf{E}vidence in knowledge-\textbf{I}ntensive \textbf{D}ialogue \textbf{G}eneration (\ourmethod{}). The labeled and unlabeled latent evidence, defined as valid knowledge fragments, supports factual responses. Our method highlights the significance of unlabeled evidence due to its ability to offer valuable insights into historical and current dialogues. To incorporate evidence into the KIDG task, we introduce an automatic evidence generation (AEG) module, which mines potential unlabeled evidence based on pre-defined criteria. The synthetic training data is adopted to train a reliable evidence indicator, which directly predicts evidence fragments. Furthermore, we incorporate an evidence-augmented generator (EAG) module with an evidence-focused attention mechanism, which ensures the effectiveness of the proposed focus-on-evidence strategy.

Our approach demonstrates the effective utilization of evidence in KIDG to generate factual responses. Experiments on the Multidoc2Dial dataset confirmed this efficacy, demonstrating a significant enhancement of around +3$\sim$+5 BLEU points over previous approaches. This significant increase underscores the advantage of utilizing evidence directly in response generation.
We also conduct human evaluations to authenticate the accuracy of the evidence labels generated by our proposed AEG module. The high scores show the effectiveness of our approach in accurately identifying potential evidence.
By directly incorporating augmented evidence into the response generation process, our approach improves the factuality beyond what previous methods have achieved.

\section{Our method}
\label{sec:method}

\begin{figure*}[htb]
\centering         
\includegraphics[width=0.85\textwidth]{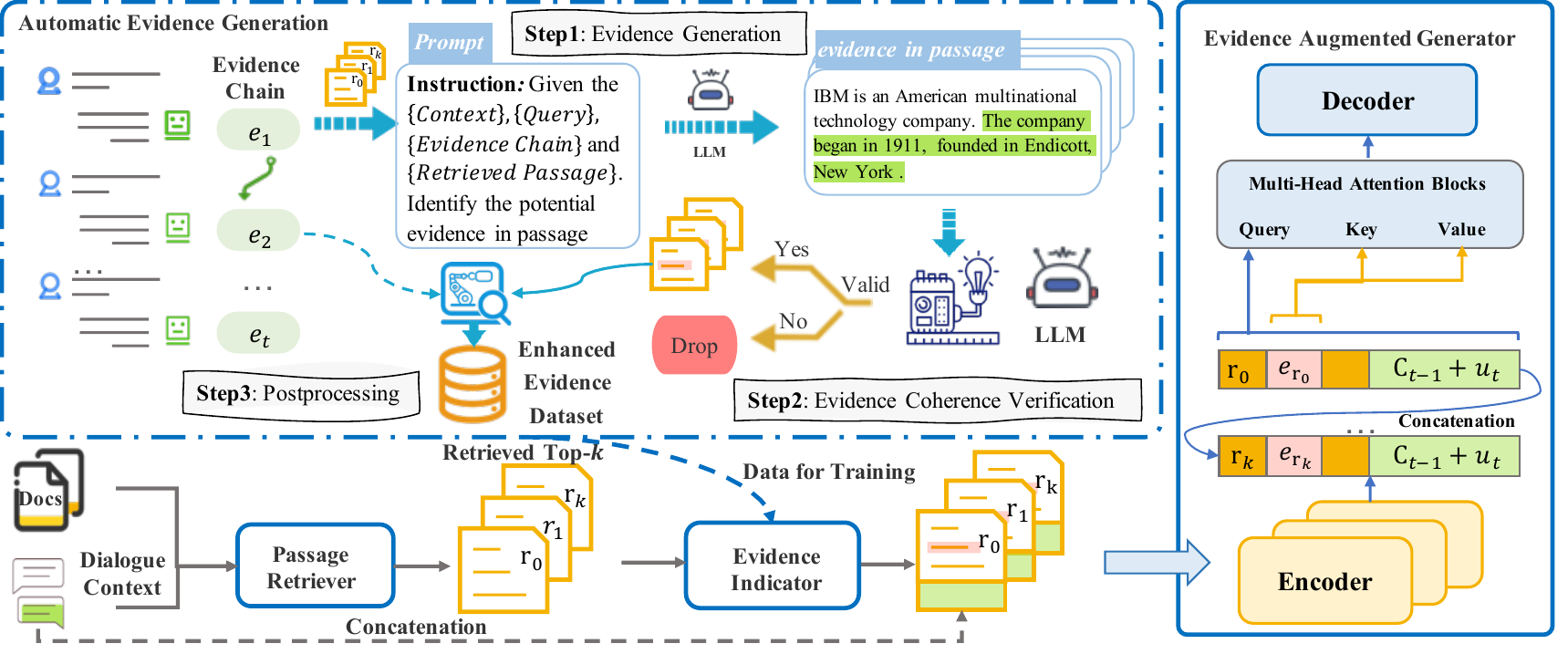}
\label{fig:model-structure}
\caption{\ourmethod{} consists of three components:(1)passage retriever, (2) evidence indicator enhanced by \textbf{A}utomatic \textbf{E}vidence \textbf{G}eneration (AEG), and (3) \textbf{E}vidence-\textbf{A}ugmented \textbf{G}enerator (EAG).} 
\vspace{-15pt}
\end{figure*}

\subsection{Overview of \ourmethod{}}
\label{ssec:overview}
Knowledge-intensive dialogue generation (\task{}) aims to enhance the factuality of conversations by introducing external knowledge resources, defined as a collection of documents $D=\{{d_{0},\dots,d_{N}}\}$. Given a dialogue context $C_{t-1}=\{u_{0}, a_{0},\dots, a_{t-1}\}$ comprising a sequence of utterances between a user and an agent, the agent needs to ground each reply on the relevant factual knowledge. Our method aims to fully utilize the evidence to generate accurate bot response $a_{t}$ based on the user query $u_{t}$. 

As shown in Fig. \ref{fig:model-structure}, we propose an automatic evidence generation (AEG) framework that leverages large language models (LLMs) to produce high-quality labeled evidence data. To incorporate the corresponding knowledge for generation, we introduce an evidence indicator trained by evidence data to predict the presence of evidence in retrieved passage candidates, where an evidence-focused attention mechanism of the evidence-augmented generator (EAG) is adopted to (generated passage fragments) to strengthen the model concentration on crucial evidence. 


\subsection{\ourmethod{}}
\label{ssec:model-details}
\subsubsection{Passage Retriever}
\label{sssec:retrieval}

Due to length limitations, we partition the documents $D$ into smaller parts known as passages $P={\{p_{0},\ldots, p_{n}\}}$.
The primary goal of the passage retriever is to identify the most relevant passage to generate the correct response.
We employ the dense passage retrieval (DPR) method to select the top-$k$ passages $R={\{r_{0},\ldots, r_{k}}\}\in{P}$ based on semantic representation similarity, considering both the conversation context $C_{t-1}$ and the current user utterance $u_{t}$ as the retrieval query.

The positive and negative examples are used to fine-tune the DPR model. Positive examples are comprised of annotated passages and query pairs. To construct negative examples, we randomly select additional user queries from the conversation and retrieve the two most highly ranked passages, along with the current query. Additionally, we incorporate the present response and dialogue history to obtain 5$\sim$15 lower-ranking results from the BM25 algorithm as supplementary negative examples.

\subsubsection{Automatic Evidence Generation}
\label{sssec:automatic-evidence-data-generation}
Evidence refers to factual excerpts from external sources that assist in generating factual responses with the given context and user query. Our automated evidence generation pipeline consists of three primary stages: \texttt{STEP-1}: generating evidence snippets from retrieved passages. \texttt{STEP-2}: verifying the consistency within the generated evidence via the chain of thought (CoT) \cite{wei2022chain}. \texttt{STEP-3}: removing redundant evidence.

\noindent\textbf{Evidence Generation}
Since relevant but unlabeled evidence still remains in the retrieved passages, we utilize the concept of the evidence chain (``a continuity of related evidence across dialogue turns on the same topic''). 
This stage of evidence generation aims to identify relevant evidence from the retrieved passages aligned with the evidence chain. The potential unlabeled evidence can be categorized into two types: current-related evidence directly related to the current utterance and historical-related evidence derived from previous dialogue turns that potentially benefit the current response.  


The evidence chain is established through the dialogue context, query, and existing evidence labels. Meanwhile, we fetch $20$ candidate passages by the dense passage retriever to facilitate evidence generation. By combining the evidence chain and passages, the prompt is fed into GPT-4 to effectively identify potential evidence, including current-related and historical-related evidence. 

\noindent\textbf{Evidence Coherence Verification}
Since each passage is predicted independently, inconsistencies across the generated evidence snippets may exist. 
To address this issue, we propose a verification process that involves prompting GPT-4 to provide contextual information regarding the evidence chain and the target response, highlighting the specific dialogue turn associated with generated evidence. Any illogical or incoherent pieces of evidence are discarded.

\noindent\textbf{Postprocessing}
The generated evidence might conflict with the original evidence labels. In such cases, we post-process the disputable labels, preserving the original labeled evidence from the initial evidence chain instead of the GPT-4 generated evidence. 

\subsubsection{Evidence Indicator}
\label{sssec:data-augmented-evidence-indicator}
To predict the evidence position $E={\{e_{r_0},\dots,e_{r_k}\}}$, we design the evidence indicator based on the RoBERTa architecture \cite{DBLP:journals/corr/abs-1907-11692}. The model takes a single retrieved passage as input, denoted as $r_{k}\in R$, along with the dialogue context $C_{t-1}$ and the current user query $u_t$. Evidence span position is calculated as follows:
\begin{MiddleEquation}
\begin{align}
\begin{split}
p_{i}^{s}=\frac{e^{s\times{h_{i}}}}{ {\textstyle \sum_{j}}s\times{h_{j}}}; \quad p_{i}^{e}=\frac{e^{e\times{h_{i}}}}{ {\textstyle \sum_{j}}e\times{h_{j}}}
\end{split}
\end{align}
\end{MiddleEquation}where $p_{i}^{s}$ is the start probability and $p_{i}^{e}$ is the end probability for each word in the passage. we employ a linear layer to process the hidden state $h_i \in \mathbb{R}^H$, representing the output hidden state from RoBERTa for the $i^{th}$ input token. The start vector $s \in \mathbb{R}^H$ and end vector $e \in \mathbb{R}^H$ are trained during this process. To determine $p_i^s$ for each word in the passage, we compute the dot product between $h_i$ and $s$ and apply a softmax operation over all words in the paragraph. The end probability $p_i^e$ is calculated using a similar method. The answer span is predicted based on the span with the highest score.



\subsubsection{Evidence-Augmented Generator}
\label{sssec:evidence-augmented-generator}


The generator consists of an encoder that encodes each retrieved passage independently with the dialogue context and jointly in the decoder. The input can be described as $x_{k}=\{\bar{S_c}, u_{0}, a_{0}, \ldots, u_{t},\bar{S_r}, r_k\} \in X$, where the $k^{th}$ retrieved passage context is denoted by $r_{k}\in R$. The special tokens $\bar{S_c}$ and $\bar{S_r}$ indicate the start position of the dialogue context and retrieved passage, respectively. 

We propose integrating the encoder hidden layer with the evidence $E$, derived from the evidence indicator discussed in Section \ref{sssec:data-augmented-evidence-indicator}. 
We introduce a novel approach to utilize binary positional embeddings to represent evidential information and indicate whether each token is within the evidence span. Additionally, we propose a cross-attention layer within the encoder output that attends explicitly to the evidence. We refer to this attention mechanism as follows:
\begin{TinyEquation}
\begin{align}
\begin{split}
H_{a} = \texttt{SF}\left(\frac{Q_{h}K_{H_{e}}^{T}}{\sqrt{d_{H}}} \right)V_{H_{e}}
\end{split}
\end{align}
\end{TinyEquation}where $\texttt{SF}(\cdot)$ denotes the softmax function, and the fused representation is defined as $H_{a}$, while $H$ represents the hidden states of the encoder output. $H_{e}$ specifically refers to the hidden states corresponding to the evidence position.

In contrast to self-attention, where the encoder hidden layer is used as the query ($Q$), key ($K$), and value ($V$), our approach employs the encoder hidden states as $Q_H$ and the evidence representations $H_{e}$ for both $K$ and $V$ during attention computation, allowing the model to prioritize context relevant to the evidence.

\section{Experiments}
\label{sec:experiments}

\subsection{Dataset and Evaluation}
\label{sec:evaluation}
Our experiments are conducted on the Multidoc2Dial \cite{feng2021multidoc2dial} benchmark, including 4,800 dialogues comprising 29,748 queries, with an average of 14 turns per conversation. The dataset provides comprehensive annotations for each dialogue turn, including speaker roles, dialogue acts, human transcripts, and evidence spans. The metrics token F1, SacreBLEU, and ROUGE-L \cite{feng2021multidoc2dial} are used for dialogue evaluation.

\subsection{Implementation Details}
We utilize automatic evidence generation (AEG) to annotate a total of 159,293 evidence labels, where the labels are defined into two categories: historical-related evidence with 66,911 labels and current-related evidence with 92,382 labels. We use the ANCE model \cite{XiongXLTLBAO21} as our dense passages retriever (DPR) method to retrieve the corresponding passages from the documents. Our approach involves training the RoBERTa model with our automatically generated dataset as the evidence indicator. We utilize the T5$_{\text{base}}$ \cite{raffel2020exploring} as the foundation of our evidence-augmented generator.

\subsection{Baselines}
\label{sec:baseline-approaches}

\noindent\textbf{RAG} \cite{lewis2020retrieval} enable end-to-end training by sharing encoder layers between the retriever and the generator. 

\noindent\textbf{FiD} \cite{izacard2021leveraging} (Fusion-in-Decoder) independently encodes retrieved passages and fuses them before the decoder.

\noindent\textbf{DIALKI} \cite{wu-etal-2021-dialki} identifies document knowledge via multi-task learning over passages and contextualized span prediction.

\noindent\textbf{EviGui-G} \cite{asai2022evidentiality} uses multi-task learning to optimize text generation by incorporating evidence prediction tasks.

\subsection{Experimental Results}
\begin{table}[htb]
\centering
\resizebox{0.95\columnwidth}{!}{
\begin{tabular}{l|ccc}
\toprule
\textbf{Method} & \textbf{F1} & \textbf{SacreBLEU} & \textbf{Rouge-L} \\ 
\midrule
RAG-BART$_{\text{large}}$    & 34.25 & 19.44 & 31.85 \\
DIALKI-BART$_{\text{large}}$ & 38.95 & 25.13 & 37.64 \\
FiD-T5$_{\text{base}}$       & 42.14 & 28.58 & 40.67 \\
EviGui-G-T5$_{\text{base}}$  & 43.14 & 30.01 & 41.33 \\
\midrule
\rowcolor[rgb]{0.85,0.92,0.83}
\ourmethod{}-T5$_{\text{base}}$  &  \textbf{46.85}  &  \textbf{33.24}   & \textbf{44.65}  \\
\quad \textit{w/o AEG}  & 44.85   &  31.69 & 43.04\\
\midrule 
\multicolumn{4}{c}{\textit{Upper Bound of our method}} 
\\ \midrule
\rowcolor[rgb]{0.81,0.89,0.95}
\ourmethod{}$_{ub}$-T5$_{\text{base}}$ & 48.56 & 34.27 & 46.75    \\
\bottomrule
\end{tabular}}
\caption{Main results of experiments on Multidoc2Dial where \ourmethod{}$_{ub}$-T5$_{\text{base}}$ uses the AEG labeled evidence instead of evidence indicator prediction }
\label{fig:main-results}
\end{table}

\noindent\textbf{Main Results} The main results of our experiments are summarized in Table \ref{fig:main-results}. Our \ourmethod{} exhibits effectiveness compared to our baseline models, with the efforts to utilize evidence. Specifically, our evidence augmentation approach outperforms EviGui-G and DIALKI regarding the exceptional evidence utilization method. Notably, the \ourmethod{} model without AEG performs less favorably than the model with all components, suggesting that the AEG module contributes positively to the capabilities of the model.

\noindent\textbf{Upper bound of \ourmethod{}} The results with blue background (upper bound of our method) explore the upper limit of the \ourmethod{} by directly inputting evidence generated by AEG into the model. The results of the \ourmethod{}$_{ub}$ model exhibit a significant improvement in overall performance compared to \ourmethod{}, indicating that the limitation lies in the evidence indicators. 
The generated response exhibits a clear demonstration of factuality through our human analysis. Notably, our evaluation metrics introduce certain limitations on the scores, as they primarily assess the closeness instead of correctness with the golden response.

\subsection{Evidence Correctness}
\label{ssec:evidence-correcness}


We manually verify the generated evidence dataset to evaluate the effectiveness of AEG. Our assessment revolves around three crucial aspects: factuality, relevance, and helpfulness.
\textbf{Factuality} means fabrications or inaccuracies in the generated data. \textbf{Relevance} assesses whether the evidence relates to the golden response. \textbf{Helpfulness} judges if the evidence provides helpful information for generating responses. Each dimension is assigned a score $s \in \{-1, 0, 1\}$, where $1$ indicates positive performance and $-1$ on the negative. 

\begin{table}[htb]
\centering
\resizebox{0.45\textwidth}{!}
{\begin{tabular}{l|ccc}
    \toprule
     &  \textbf{Factuality} & \textbf{Relevance} & \textbf{helpfulness} \\
    \midrule
    \textbf{Current-Related} & 0.92 & 0.83 & 0.72 \\
    \textbf{Historical-Related} & 0.84 & 0.61 & 0.54  \\
    \bottomrule
\end{tabular}}
\caption{Human assessment evidence dataset on sampled 300 cases comprising historical and current-related evidence.}
\label{table:human_evaluation}
\vspace{-5pt}
\end{table}

The AEG manual evaluation results affirm its overall quality in Table \ref{table:human_evaluation}. Our analysis reveals that the performance of historical-related evidence data is comparatively lower, attributed to the challenge of determining the relevance of information across multiple dialogue turns. On the other hand, the high quality of current-related evidence data attests to the effectiveness of our data framework in mining potential 
factual evidence from previously unannotated retrieved passages.

\section{Conclusion}
\label{sec:conclusion}

In this paper, our work focuses on the importance of evidence in knowledge-intensive dialogue generation (\task{}). To effectively incorporate the evidence into \task{}, we propose a novel approach called \ourmethod{}. Our method utilizes automatic evidence generation (AEG) to identify potential evidence fragments. These factual evidence fragments are integrated into an evidence-augmented generator (EAG) utilizing an evidence-focused attention mechanism. Experimental results demonstrate that our method maximizes the potential of evidence in knowledge-intensive dialogue generation.


\vfill\pagebreak
\bibliographystyle{IEEEbib}
\bibliography{refs}

\begin{thebibliography}{10}

\bibitem{dinan2018wizard}
Emily Dinan, Stephen Roller, Kurt Shuster, Angela Fan, Michael Auli, and Jason Weston,
\newblock ``Wizard of wikipedia: Knowledge-powered conversational agents,''
\newblock in {\em ICRL}, 2018.

\bibitem{feng2021multidoc2dial}
Song Feng, Siva~Sankalp Patel, Hui Wan, and Sachindra Joshi,
\newblock ``Multidoc2dial: Modeling dialogues grounded in multiple documents,''
\newblock in {\em EMNLP}, 2021, pp. 6162--6176.

\bibitem{li-etal-2022-knowledge}
Yu~Li, Baolin Peng, Yelong Shen, Yi~Mao, Lars Liden, Zhou Yu, and Jianfeng Gao,
\newblock ``Knowledge-grounded dialogue generation with a unified knowledge representation,''
\newblock in {\em NAACL}, 2022, pp. 206--218.

\bibitem{li-etal-2022-enhancing-knowledge}
Sha Li, Mahdi Namazifar, Di~Jin, Mohit Bansal, Heng Ji, Yang Liu, and Dilek Hakkani-Tur,
\newblock ``Enhancing knowledge selection for grounded dialogues via document semantic graphs,''
\newblock in {\em NAACL}, 2022, pp. 2810--2823.

\bibitem{kinet}
Jiaqi Bai, Ze~Yang, Jian Yang, Hongcheng Guo, and Zhoujun Li,
\newblock ``Kinet: Incorporating relevant facts into knowledge-grounded dialog generation,''
\newblock {\em TASLP}, vol. 31, pp. 1213--1222, 2023.

\bibitem{griprank}
Jiaqi Bai, Hongcheng Guo, Jiaheng Liu, Jian Yang, Xinnian Liang, Zhao Yan, and Zhoujun Li,
\newblock ``Griprank: Bridging the gap between retrieval and generation via the generative knowledge improved passage ranking,''
\newblock {\em CoRR}, vol. abs/2305.18144, 2023.

\bibitem{voorhees-tice-2000-trec}
Ellen~M. Voorhees and Dawn~M. Tice,
\newblock ``The {TREC}-8 question answering track,''
\newblock in {\em LREC}, 2000.

\bibitem{joshi-etal-2017-triviaqa}
Mandar Joshi, Eunsol Choi, Daniel Weld, and Luke Zettlemoyer,
\newblock ``{T}rivia{QA}: A large scale distantly supervised challenge dataset for reading comprehension,''
\newblock in {\em ACL}, 2017, pp. 1601--1611.

\bibitem{SachanRHDY21}
Devendra~Singh Sachan, Siva Reddy, William~L. Hamilton, Chris Dyer, and Dani Yogatama,
\newblock ``End-to-end training of multi-document reader and retriever for open-domain question answering,''
\newblock in {\em NeurIPS}, 2021, pp. 25968--25981.

\bibitem{chen-etal-2017-reading}
Danqi Chen, Adam Fisch, Jason Weston, and Antoine Bordes,
\newblock ``Reading {W}ikipedia to answer open-domain questions,''
\newblock in {\em ACL}, 2017, pp. 1870--1879.

\bibitem{chen-etal-2020-bridging}
Xiuyi Chen, Fandong Meng, Peng Li, Feilong Chen, Shuang Xu, Bo~Xu, and Jie Zhou,
\newblock ``Bridging the gap between prior and posterior knowledge selection for knowledge-grounded dialogue generation,''
\newblock in {\em EMNLP}, 2020, pp. 3426--3437.

\bibitem{lewis2020retrieval}
Patrick S.~H. Lewis, Ethan Perez, Aleksandra Piktus, Fabio Petroni, Vladimir Karpukhin, Naman Goyal, Heinrich K{\"{u}}ttler, Mike Lewis, Wen{-}tau Yih, Tim Rockt{\"{a}}schel, Sebastian Riedel, and Douwe Kiela,
\newblock ``Retrieval-augmented generation for knowledge-intensive {NLP} tasks,''
\newblock in {\em NeurIPS}, 2020.

\bibitem{karpukhin2020dense}
Vladimir Karpukhin, Barlas Oguz, Sewon Min, Patrick Lewis, Ledell Wu, Sergey Edunov, Danqi Chen, and Wen-tau Yih,
\newblock ``Dense passage retrieval for open-domain question answering,''
\newblock in {\em EMNLP}, 2020.

\bibitem{alm}
Jian Yang, Shuming Ma, Dongdong Zhang, Shuangzhi Wu, Zhoujun Li, and Ming Zhou,
\newblock ``Alternating language modeling for cross-lingual pre-training,''
\newblock in {\em AAAI}. 2020, pp. 9386--9393, {AAAI} Press.

\bibitem{ganlm}
Jian Yang, Shuming Ma, Li~Dong, Shaohan Huang, Haoyang Huang, Yuwei Yin, Dongdong Zhang, Liqun Yang, Furu Wei, and Zhoujun Li,
\newblock ``Ganlm: Encoder-decoder pre-training with an auxiliary discriminator,''
\newblock in {\em ACL}, 2023, pp. 9394--9412.

\bibitem{izacard2021leveraging}
Gautier Izacard and {\'E}douard Grave,
\newblock ``Leveraging passage retrieval with generative models for open domain question answering,''
\newblock in {\em EACL}, 2021, pp. 874--880.

\bibitem{glass2022re2g}
Michael Glass, Gaetano Rossiello, Md~Faisal~Mahbub Chowdhury, Ankita~Rajaram Naik, Pengshan Cai, and Alfio Gliozzo,
\newblock ``Re2g: Retrieve, rerank, generate,''
\newblock in {\em NAACL}, 2022.

\bibitem{geigle2022retrieve}
Gregor Geigle, Jonas Pfeiffer, Nils Reimers, Ivan Vuli{\'c}, and Iryna Gurevych,
\newblock ``Retrieve fast, rerank smart: Cooperative and joint approaches for improved cross-modal retrieval,''
\newblock {\em TACL}, pp. 503--521, 2022.

\bibitem{izacard2021distilling}
Gautier Izacard and Edouard Grave,
\newblock ``Distilling knowledge from reader to retriever for question answering,''
\newblock in {\em ICLR}, 2021.

\bibitem{zheng-etal-2023-contextual}
Wen Zheng, Natasa Milic-Frayling, and Ke~Zhou,
\newblock ``Contextual knowledge learning for dialogue generation,''
\newblock in {\em ACL}, 2023, pp. 7822--7839.

\bibitem{wu-etal-2021-dialki}
Zeqiu Wu, Bo-Ru Lu, Hannaneh Hajishirzi, and Mari Ostendorf,
\newblock ``{DIALKI}: Knowledge identification in conversational systems through dialogue-document contextualization,''
\newblock in {\em EMNLP}, 2021, pp. 1852--1863.

\bibitem{asai2022evidentiality}
Akari Asai, Matt Gardner, and Hannaneh Hajishirzi,
\newblock ``Evidentiality-guided generation for knowledge-intensive nlp tasks,''
\newblock in {\em NAACL}, 2022, pp. 2226--2243.

\bibitem{zhang2023coarse}
Yeqin Zhang, Haomin Fu, Cheng Fu, Haiyang Yu, Yongbin Li, and Cam-Tu Nguyen,
\newblock ``Coarse-to-fine knowledge selection for document grounded dialogs,''
\newblock in {\em ICASSP}, 2023, pp. 1--5.

\bibitem{wei2022chain}
Jason Wei, Xuezhi Wang, Dale Schuurmans, Maarten Bosma, Fei Xia, Ed~Chi, Quoc~V Le, Denny Zhou, et~al.,
\newblock ``Chain-of-thought prompting elicits reasoning in large language models,''
\newblock {\em NeurIPS}, pp. 24824--24837, 2022.

\bibitem{DBLP:journals/corr/abs-1907-11692}
Yinhan Liu, Myle Ott, Naman Goyal, Jingfei Du, Mandar Joshi, Danqi Chen, Omer Levy, Mike Lewis, Luke Zettlemoyer, and Veselin Stoyanov,
\newblock ``Roberta: {A} robustly optimized {BERT} pretraining approach,''
\newblock {\em CoRR}, 2019.

\bibitem{XiongXLTLBAO21}
Lee Xiong, Chenyan Xiong, Ye~Li, Kwok{-}Fung Tang, Jialin Liu, Paul~N. Bennett, Junaid Ahmed, and Arnold Overwijk,
\newblock ``Approximate nearest neighbor negative contrastive learning for dense text retrieval,''
\newblock in {\em ICLR}, 2021.

\bibitem{raffel2020exploring}
Colin Raffel, Noam Shazeer, Adam Roberts, Katherine Lee, Sharan Narang, Michael Matena, Yanqi Zhou, Wei Li, and Peter~J Liu,
\newblock ``Exploring the limits of transfer learning with a unified text-to-text transformer,''
\newblock {\em JMLR}, pp. 5485--5551, 2020.

\end{thebibliography}

\end{document}